\documentclass{ws-ijns}

\usepackage[english]{babel}
\usepackage[super,sort,compress,comma]{natbib}

\usepackage{url}
\usepackage{hyperref}
\usepackage{graphicx}
\usepackage{subfigure}
\usepackage{algorithm}
\usepackage{algpseudocode}
\usepackage{amsmath}
\usepackage{graphics}
\usepackage{epsfig}
\usepackage{verbatimbox}
\usepackage{multirow}
\usepackage{makecell}
\usepackage{color}

\usepackage{threeparttable}

\begin{document}


\markboth{Zhang et al.}{Learning High-quality Proposals for Acne Detection}

\title{LEARNING HIGH-QUALITY PROPOSALS FOR ACNE DETECTION}


\author{
	JIANWEI ZHANG, LEI ZHANG\footnote{Corresponding author}, JUNYOU WANG, XIN WEI
}

\address{College of Computer Science, Sichuan University, Section 4, Southern 1st Ring Rd,\\
	Wuhou District, Chengdu, Sichuan 610065, China\\
	zhangjianwei@stu.scu.edu.cn\\
	leizhang@scu.edu.cn\\
	wangjunyou@stu.scu.edu.cn\\
	weixin@stu.scu.edu.cn\\
}

\author{
	JIAQI LI, XIAN JIANG and DANDU
}

\address{Department of Dermatology, West China Hospital, Sichuan University, No.37 Guoxue Alley,\\
	Wuhou District, Chengdu, Sichuan 610041, China\\
	lijiaqicd@gmail.com\\
	youradrian@outlook.com\\
	dudan.meishan@foxmail.com\\
}

\maketitle

\begin{abstract}
	Acne detection is crucial for interpretative diagnosis and precise treatment of skin disease. The arbitrary boundary and small size of acne lesions lead to a significant number of poor-quality proposals in two-stage detection. In this paper, we propose a novel head structure for Region Proposal Network to improve the proposals' quality in two ways. At first, a Spatial Aware Double Head(SADH) structure is proposed to disentangle the representation learning for classification and localization from two different spatial perspectives. The proposed SADH ensures a steeper classification confidence gradient and suppresses the proposals having low intersection-over-union(IoU) with the matched ground truth. Then, we propose a Normalized Wasserstein Distance prediction branch to improve the correlation between the proposals' classification scores and IoUs. In addition, to facilitate further research on acne detection, we construct a new dataset named AcneSCU, with high-resolution imageries, precise annotations, and fine-grained lesion categories. Extensive experiments are conducted on both AcneSCU and the public dataset ACNE04, and the results demonstrate the proposed method could improve the proposals' quality, consistently outperforming state-of-the-art approaches. Code and the collected dataset are available in \url{https://github.com/pingguokiller/acnedetection}.
\end{abstract}


\begin{multicols}{2}
	\section{Introduction}
	\label{sec:introduction}
	Acne vulgaris, well-known as acne, is one of the most popular skin disorders in dermatology~\cite{bernardis2020development, wu2019joint}. Delay of acne treatment would lead to not only physical disfigurements such as scars and pigmentation, but also psychosocial impacts such as social isolation, depression, and suicidality~\cite{barbieri2021patient}. Counting skin lesions of different categories is the first and most crucial step of routine clinical acne treatment. Nevertheless, this step is usually manually performed by dermatologists~\cite{kittigul2016automatic, maroni2017automated}, which is time-consuming and highly dependent on medical expertise. 
	
	With the rapid development of deep learning, automated acne detection recently drew a growing interest~\cite{rashataprucksa2020acne,min2021acnet}. Rashataprucksa et al.~\cite{rashataprucksa2020acne} proposed to handle acne detection with deep learning, and achieved faster and better performance than traditional methods. Min et al.~\cite{min2021acnet} proposed a deep learning based method named ACNet to address several problems on acne detection, such as inconsistent illumination, various lesion scales, and high-density distribution. In summary, previous approaches mostly focus on two-stage detection, which treats the detection as a coarse-to-fine process.
	
	\begin{figurehere}
		\centerline{\includegraphics[width=0.48\textwidth]{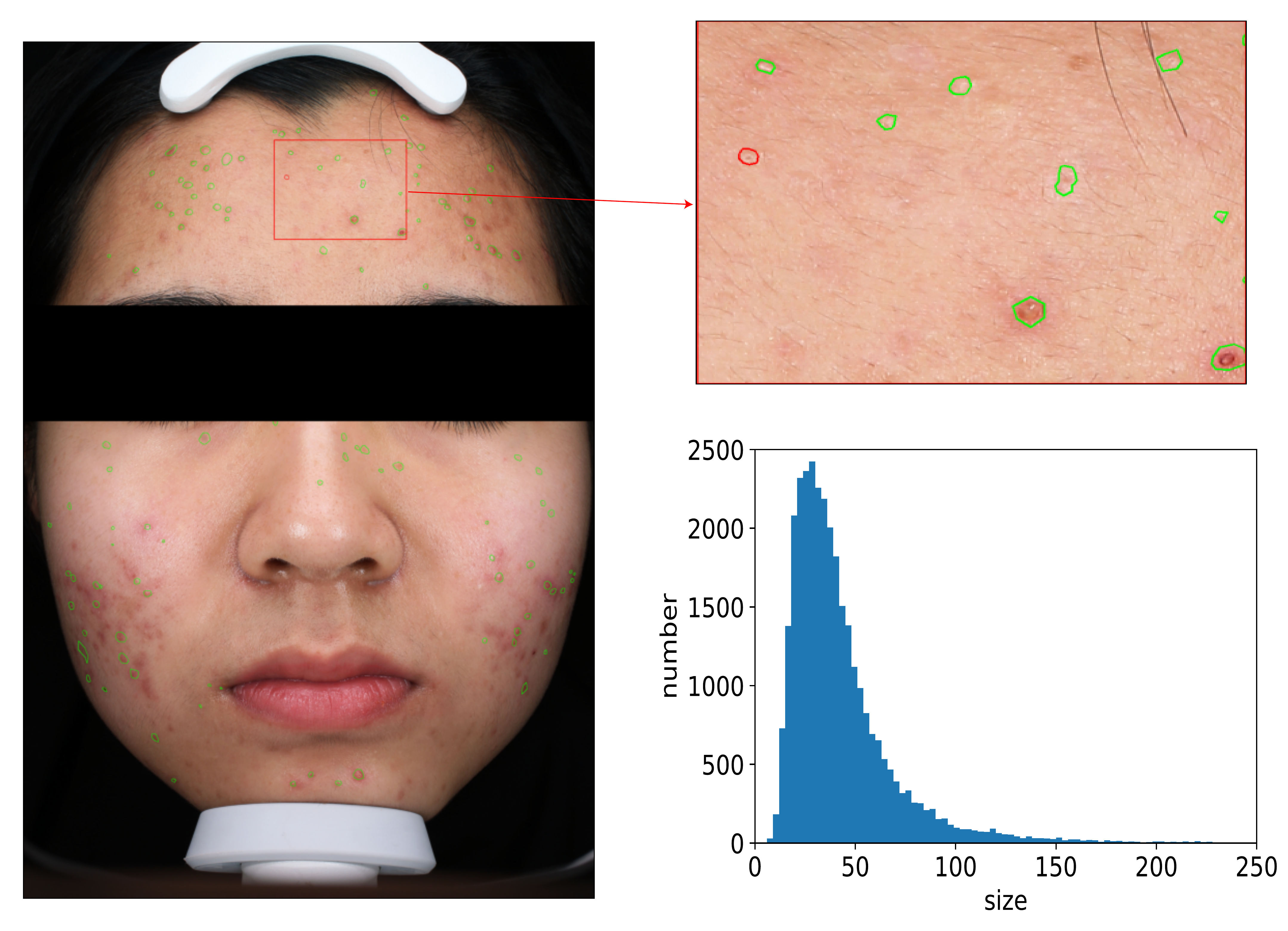}}
		\caption{Left: an example from AcneSCU, where the image has a size of $3456\times5184$. Top right: a partial region with several lesions from the left example, where the lesion with red boundary is an open comedone with an area about $28\times28$ pixels; Bottom right: the lesion size distribution of AcneSCU. Please note that this paper defines the lesion size as the square root of the bounding box area.}
		\label{fig_dataset}
	\end{figurehere}
	
	In the first stage of two-stage detection methods, a Region Proposal Network(RPN) is usually utilized to generate a number of coarse object proposals~\cite{ren2015faster}. RPN consists of two tasks: the classification task to predict a score to indicate how likely a proposal contains an object and the localization task to predict the expected object's bounding box. Therefore, every proposal is essentially a bounding box with a classification score. Especially, when training the classification task, label assignment selects positive and negative samples according to the anchors' IoUs with their corresponding ground truth bounding boxes through an IoU threshold, which is called \textbf{label threshold} for simplicity in this paper. In the second stage, the proposals with top classification scores would be inputted to RCNN~\cite{girshick2014rich} and refined for more precise classification and localization, while the others would be filtered through a classification score threshold, which is called the \textbf{proposal threshold} for simplicity in this paper. For two-stage detection methods, their performance highly depends on a high Average Recall(AR) of proposals generated by RPN~\cite{hosang2015makes,zou2019object}.
	
	\begin{figurehere}
		\centerline{\includegraphics[width=0.48\textwidth]{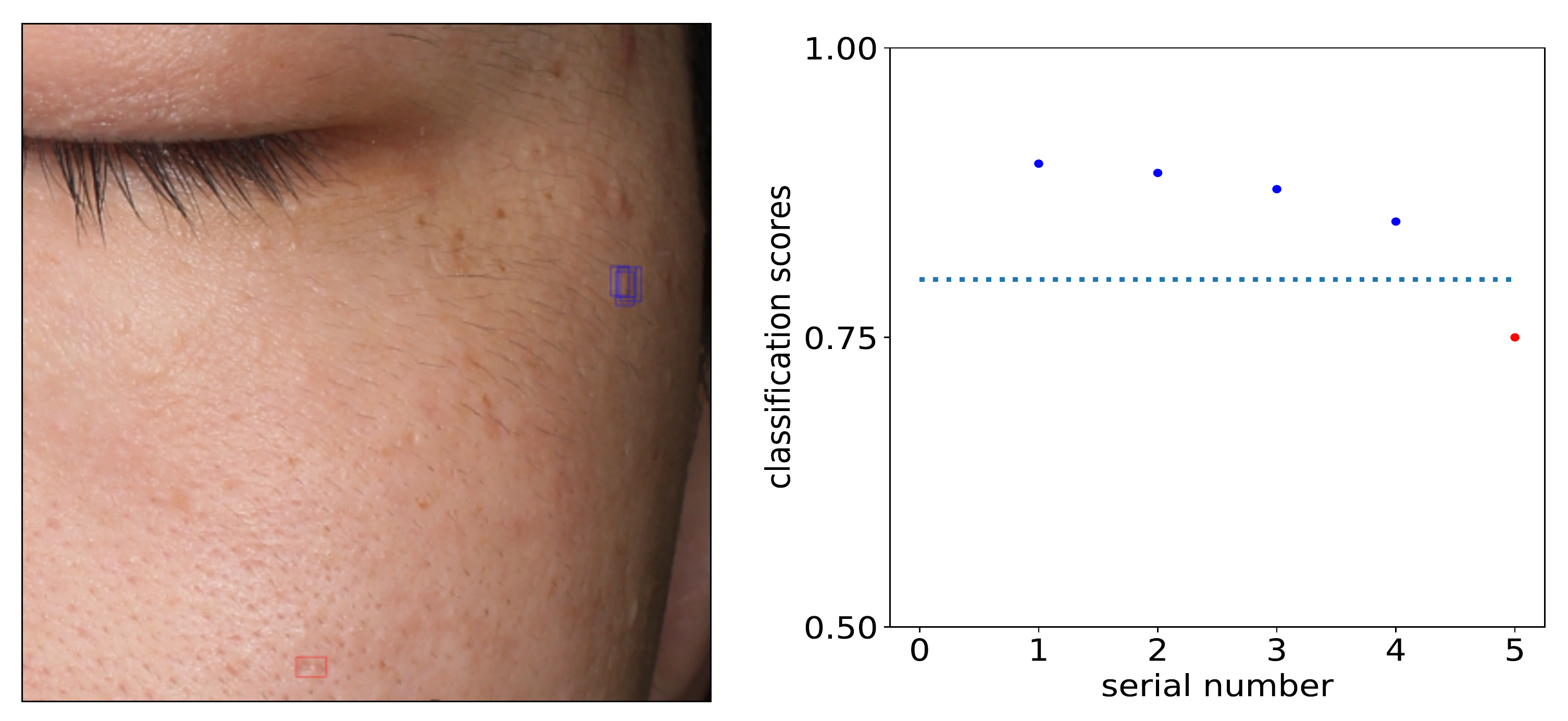}}
		\caption{Left: an illustrating example about the RPN proposals, where the blue and red bounding boxes are the proposals of easy sample and hard sample, respectively. Right: the corresponding classification scores of the proposals in the left sub-figure. It could be easily found the classification score gradient curve is very flat. The dotted line indicates the proposal threshold that filters the proposal of the hard sample.}
		\label{fig_hard_easy}
	\end{figurehere}
	
	However, acne detection suffers the arbitrary boundary and small average size of lesions, shown in Fig.~\ref{fig_dataset}, aggravating the low quality of RPN proposals. Specifically, a low AR of proposals is always observed in RPN, suppressing the performance of two-stage detection methods. In this paper, we argue the pool quality of proposals is caused by two factors. On the one hand, as shown in Fig.~\ref{fig_hard_easy}, the classification scores of the proposals for easy samples are usually very high and pretty close, while that for hard samples are usually very low and filtered. As a result, RPN tends to output redundant proposals for easy samples and ignore the hard samples, leading to a low AR. In this paper, we refer to this phenomenon as \textbf{flat classification confidence gradient} problem. On the other hand, label assignment in RPN utilize a label threshold to split the anchors into the positive and negative samples. Two samples around the two sides of the label threshold may have very close IoUs, but their classification scores are very different. For example, if the label threshold is 0.5, the IoUs of two anchors are 0.49 and 0.51, and their classification labels in RPN would be 0 and 1, respectively. As a result, the unsmooth label assignment leads to a low correlation between the classification scores and the proposals' IoUs. 
	
	In this paper, we propose a novel detection head structure named Spatial Aware Double Head(SADH) to address the flat classification confidence gradient problem. Specifically, the representation learning for classification and localization are disentangled by two different convolution layers. For the classification task, as shown in Fig.~\ref{fig_SADH}, a \emph{$1\times1$ convolution} is utilized to constrain the classification scores of the proposals at adjacent positions to be more independent. For the localization task, a \emph{$3\times3$ convolution} is utilized to constrain the classification scores at adjacent positions to be related. To address the low correlation between the classification scores and the proposals' IoUs, we propose to predict Normalized Wasserstein Distance(NWD), which is a more appropriate localization confidence metric for small objects~\cite{wang2021normalized} and learned by a novel Soft-style Binary Cross Entropy(SBCE) loss. Then, the NWDs are utilized as localization confidence to rectify the original classification scores. In addition, to comprehensively evaluate the proposed method and facilitate further research on acne detection, a new dataset named AcneSCU, containing 276 images and 31777 annotations with 10 lesions categories, is collected. Experimental results on both AcneSCU and public dataset ACNE04 show that the proposed method outperforms state-of-the-art methods. Our contributions can be summarized as follows:
	
	\begin{enumerate}
		\item In order to support and facilitate sophisticated research on acne detection, a new dataset called AcneSCU is collected. Compared to previous datasets for acne detection, AcneSCU has higher-resolution and normalized imageries, fine-grained acne categories, and more precise annotations.
		\item The high-resolution images of AcneSCU calls for cropping the images at first, but the unavoidable partial acne lesions on edge would introduce inaccurate training information. To bypass this issue, we proposed a simple but efficient data preprocessing method named masked crop by masking all the partial lesions. 	
		\item A new head structure named SADH is proposed to address the flat classification confidence gradient problem. The representation learning for classification and localization is disentangled by two different convolution layers, generating a steeper classification confidence gradient and suppressing the easy samples' proposals which have low IoUs with the matched ground truths.
		\item A NWD prediction branch along with an SBCE loss is proposed for localization confidence prediction. It has been demonstrated the classification scores rectified by NWDs could improve not only the correlation between classification scores and the proposals' IoUs but also the overall detection performance.
	\end{enumerate}

	\begin{figurehere}
	\begin{center}
		\subfigure[origin RPN Head]{
			\includegraphics[height=0.2\textwidth, width=0.46\textwidth]{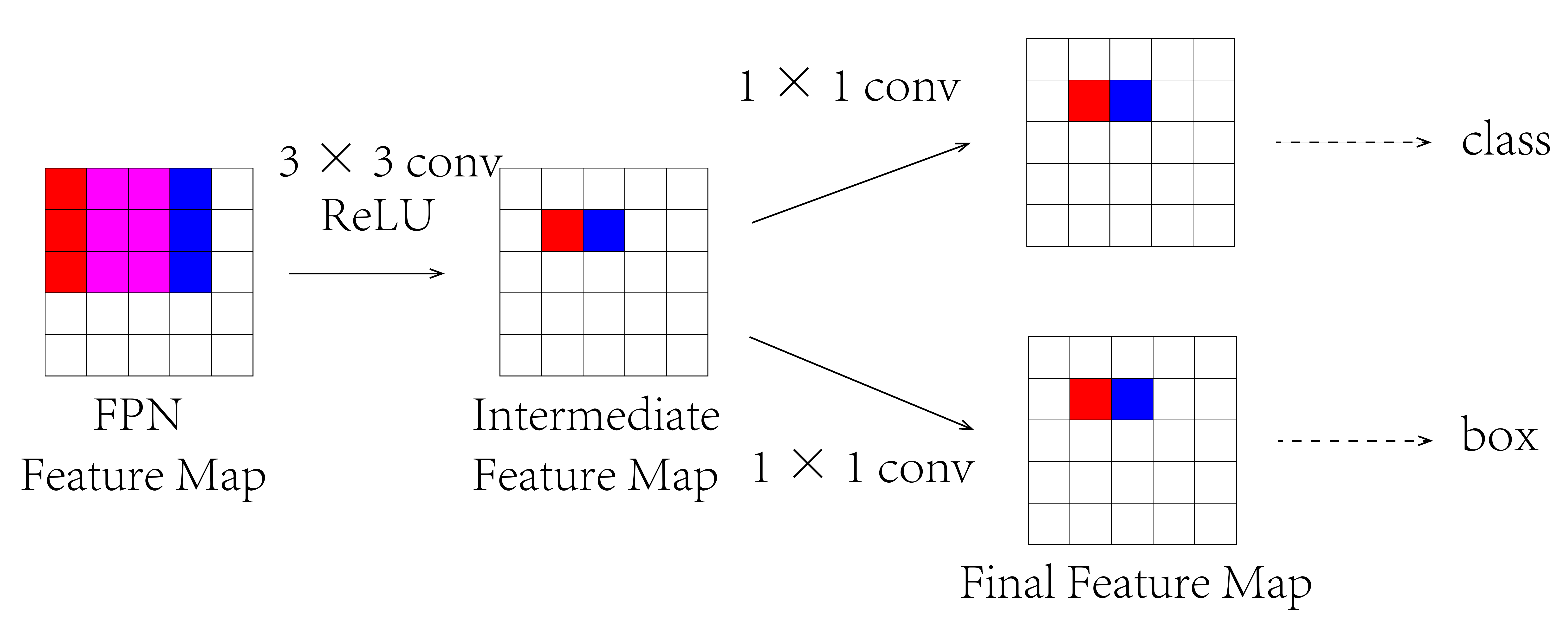}		
			\label{fig_RPN}
		}
		\subfigure[SADH and the NWD precition branch]{
			\includegraphics[height=0.34\textwidth, width=0.46\textwidth]{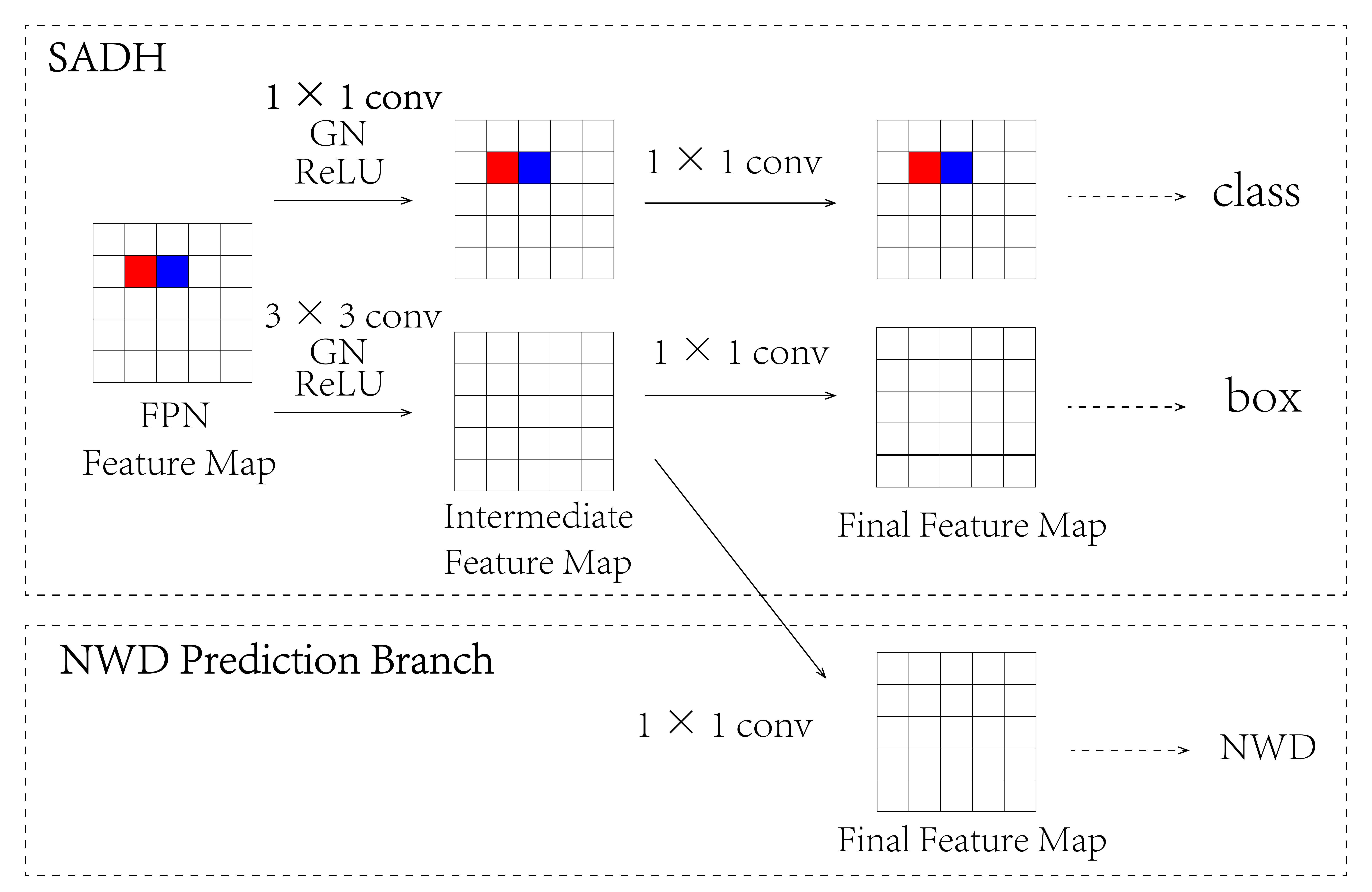}			
			\label{fig_SADH}
		}		
		\caption{(a) The RPN head. The red element on the final feature map is calculated by red and purple elements on FPN feature map, while the blue element on the final feature map is calculated by blue and purple elements. (b) The proposed SADH and NWD prediction branch. The inputs of classification scores at adjacent positions on the final feature map are independent, unlike that in RPN have same inputted purple elements, leading to a steeper classification confidence gradient.}
		\label{fig_head}
	\end{center}
	\end{figurehere} 
	
	\section{Related Work}
	\subsection{Detection Heads}
	Detection heads have been widely used in object detection and finely designed for different tasks. RPN~\cite{ren2015faster}, widely adopted in two-stage detection methods, is essentially a kind of fully-convolutional network(FCN)~\cite{long2015fully} to generate proposals at each feature map position. As shown in Fig.~\ref{fig_RPN}, RPN consists of two branches for classification and localization tasks, respectively. Law et al.~\cite{law2018cornernet} proposed a Cornernet to predict an object's top-left corner and bottom-right corner points. Duan et al.~\cite{duan2019centernet} further extended Cornernet by predicting the center point of an object. Ma et al.~\cite{ma2021anchor} proposed to predict a matching degree score between the predicted bounding box and the ground truth as well as the IoU. Sun et al.~\cite{sun2020exploring} proposed to a universal relation exploring scheme to explore the relations of different entity level, such as objects, superpixels and pixels. Mask R-CNN~\cite{he2017mask} extended Faster R-CNN by adding an instance segmentation head. Wu et al.~\cite{wu2020rethinking} conducted a thorough comparison between the fully connected head(\emph{fc-head}) and the convolution head(\emph{conv-head}) in R-CNN, and found an interesting phenomenon: \emph{fc-head} is more suitable for the classification task, while \emph{conv-head} is more suitable for the localization task.
	
	The proposed SADH structure is related to the thought-provoking work~\cite{wu2020rethinking}. We both introduce a linear-style layer for the classification task. The difference is that \cite{wu2020rethinking} performs a \emph{fc-head} on the features from RoIAlign, while the proposed method introduces a computationally efficient \emph{1$\times$1 convolution} in RPN, see Fig.~\ref{fig_SADH} for example. However, our motivations are thoroughly different. Wu et al. are motivated by the fact \emph{fc-head} has more spatial sensitivity than \emph{conv-head}, while the proposed SADH is designed to improve the steepness of the classification confidence gradient. 
	
	\subsection{Localization Confidence Prediction}
	Recently, localization confidence prediction has also drawn a growing attention. FCOS~\cite{tian2019fcos} introduced a center-ness branch to predict the distance between a pixel and the center of its corresponding ground truth. Jiang et al.~\cite{jiang2018acquisition} proposed to predict the intersection-over-union(IoU) between the bounding box and the matched ground truth through an IoU-Net. Ma et al.~\cite{ma2021anchor} proposed to utilize the matching degree score of the corner points between the predicted bounding box and the ground truth to indicate the localization confidence. Mask Scoring R-CNN~\cite{huang2019mask} proposed to predict mask IoU for better segmentation performance based on IoU-Net and Mask R-CNN. 
	
	FCOS, IoU-Net, and Mask Scoring R-CNN are all related to the proposed method. FCOS predicts a center-ness at each feature map position to down-weight the low-quality detections based on the center prior rule~\cite{zhu2020autoassign}. IoU-Net predicts the IoU and utilizes it in the non-maximum suppression(NMS) by preserving accurately localized bounding boxes. Mask Scoring R-CNN predicts a mask IoU for the instance segmentation task. In this paper, we also add a new branch to predict localization confidence. The difference is that we adopt the NWD as the metric to evaluate localization confidence, which is more suitable for detecting small acne lesions~\cite{wang2021normalized}. In addition, the NWD is learned by a novel SBCE loss and directly utilized to rectify the original classification scores.
	
	\section{Method}
	In this section, we first describe the data acquisition and the proposed data preprocessing method in Subsection~\ref{subsec_dataacqui}. Then, the details of the proposed Spatial Aware Double Head structure and the NWD prediction branch are introduced in Subsection~\ref{subsec_SADH} and Subsection~\ref{subsec_nwdpre}, respectively. Lastly, the training and inferring of the proposed method are described in Subsection~\ref{subsec_traininfer}.
	
	\subsection{Data acquisition and preprocessing}\label{subsec_dataacqui}
	\subsubsection{AcneSCU Dataset}
	To comprehensively evaluate the proposed method and facilitate further research on acne detection, we construct a new dataset called AcneSCU, which has higher-resolution and more normalized imageries, more fine-grained acne categories, and more precise annotations than previous acne dataset~\cite{wu2019joint}. AcneSCU consists of 276 facial images shot by VISIA complexion analysis system with 31777 instance segmentation annotations of 10 lesion categories, namely \emph{open comedone}, \emph{closed comedone}, \emph{papule}, \emph{pustule}, \emph{nodule}, \emph{atrophic scar}, \emph{hypertrophic scar}, \emph{melasma}, \emph{nevus} and \emph{other}. Specifically, the lesions, which are difficult to identify, are labeled as the \emph{other} class. The resolution of images varies from $3128\times4171$ to $3456\times5184$. In order to ensure the label accuracy, the annotations are labeled by 6 dermatologists and verified by 6 times. An example of AcneSCU and the distribution of lesion size are shown in Fig.~\ref{fig_dataset}. 
	
	\subsubsection{Masked Crop} 
	As most images of AcneSCU are of high resolution around $3500\times5000$, directly loading the images into memory is computationally costly. It's natural to crop each image into several sub-images. However, a large number of lesions per image makes it difficult to ensure each lesion to be cropped as an entirety. Whether keeping or ignoring the annotations of the cropped partial lesions would lead to a large number of inaccurate annotations and inferior performance. In this paper, we propose a simple but efficient way named \textbf{masked crop} by masking all the partial lesions. Each sub-image is cropped with a size of $1024\times1024$. To ensure the whole area of the image to participate in training, a total number of ${\lceil \frac{W}{1024} \rceil}\times{\lceil \frac{H}{1024} \rceil}$ sub-images are cropped out, where $W$ and $H$ denotes the width and height of the original image, respectively. Each whole image is first cropped along the horizontal direction with equal overlaps, and then cropped along the vertical direction in the same way. As there exists an overlap between the adjacent sub-images, few lesion annotations are omitted. An example of the masked crop is shown in Fig.~\ref{fig_maskedcrop}.
	
	\begin{figurehere}
		\centering{\includegraphics[width=0.46\textwidth]{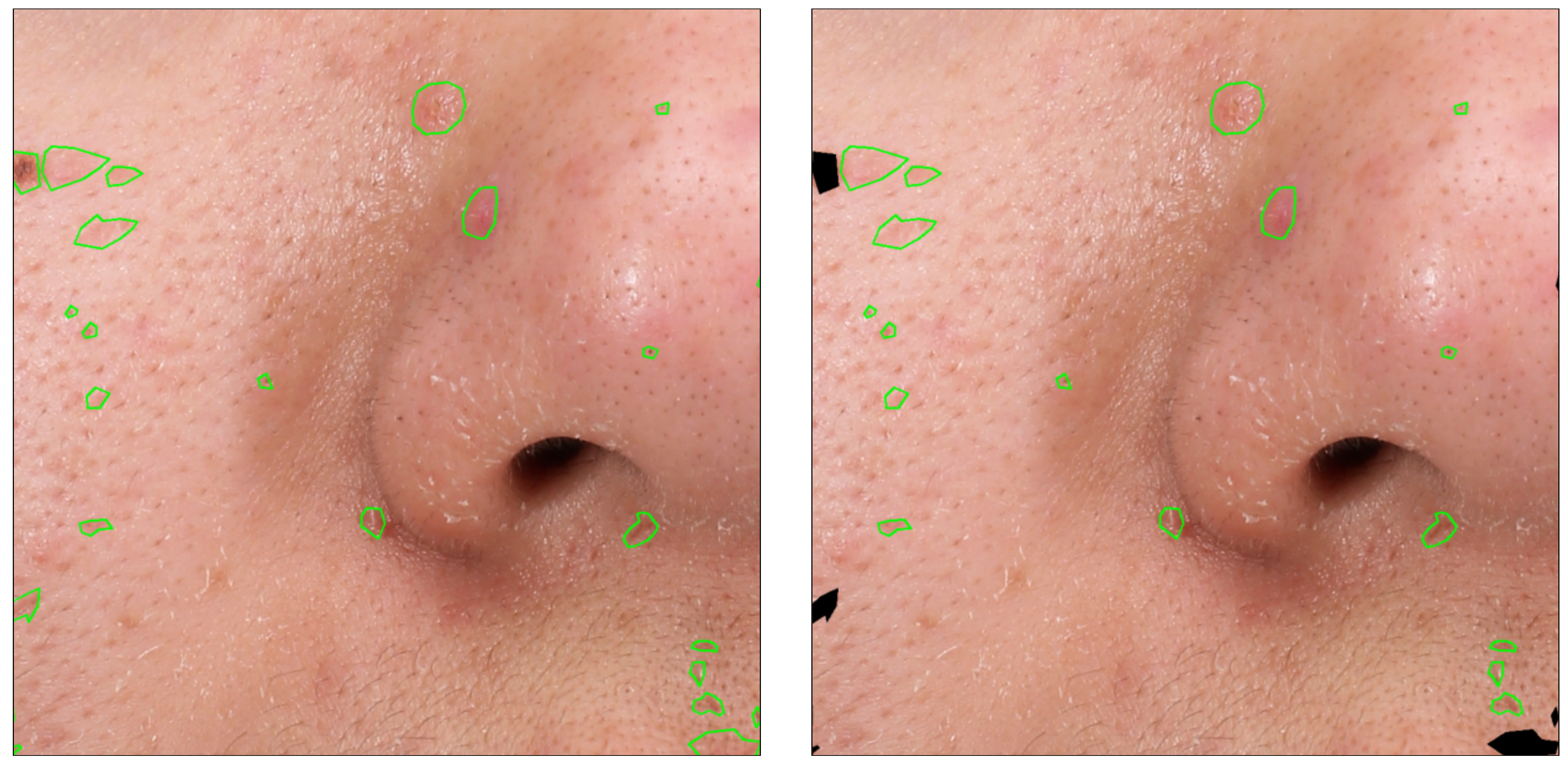}}
		\caption{Left: the vanilla crop with keeping the annotations of the partial lesions; Right: the masked crop.}
		\label{fig_maskedcrop}
	\end{figurehere}
	
	\subsection{Spatial Aware Double Head}\label{subsec_SADH}
	As shown in Fig.~\ref{fig_RPN}, vanilla RPN performs a \emph{$3\times3$ convolution} on Feature Pyramid Networks(FPN) feature map, where the same six elements would be inputted to the calculation of adjacent elements on the following intermediate and final feature maps. As a result, the classification scores at adjacent positions on the final feature map are pretty close, leading to a flat classification confidence gradient. We argue that adjacent elements on the final feature maps for the classification task should be spatially independent for a steeper classification confidence gradient, while that for the localization task should be spatially correlated for similar regression results. From the perspective of spatial perception, we propose a Spatial Aware Double Head to improve the drawback of flat classification confidence gradient in the original RPN. As shown at the top of Fig.~\ref{fig_SADH}, the intermediate feature maps for the two tasks of RPN are disentangled into two parts, which are computed with a \emph{$1\times1$} and \emph{$3\times3$ convolution} on the FPN feature map respectively. 
	
	For the classification task, the \emph{$1\times1$ convolution} is crucial to constrain the adjacent elements of final feature maps spatially independent. The red and blue elements on the final feature map are only related to the red and blue on the FPN feature map, respectively. As a result, the classification scores of proposals at adjacent positions are not necessarily too close, thus making the classification scores more discriminative. For the localization task, the intermediate feature map is computed by \emph{$3\times3$ convolution}, which is the same as RPN. As shown in Fig.~\ref{fig_RPN}, the six purple elements in the FPN feature map all participate in the computation of the red and blue elementsfine-grained. The spatial correlation benefits the regression results of proposals at adjacent positions. Moreover, following \cite{tian2019fcos}, a group normalization(GN) is added before the ReLU layer to make the training more stable.
	
	\subsection{NWD Prediction Branch}\label{subsec_nwdpre}
	Label assignment in RPN selects positive and negative samples according to a label threshold and the anchors' IoUs with the matched ground truths. The anchors with IoUs under the label threshold are regarded as negative samples, and the corresponding proposals are expected to yield extremely low classification scores, leading to a low correlation between the classification scores and the proposals' IoUs. Jiang et al.~\cite{jiang2018acquisition} proposed to predict IoU to alleviate the issue caused by the low correlation. However, Wang et al.~\cite{wang2021normalized} has demonstrated that NWD was a more suitable metric than IoU to evaluate the similarity of two bounding boxes in small object detection. In this paper, a new NWD prediction branch is proposed by performing a \emph{$1\times1$ convolution} on the intermediate feature maps for the localization task, as shown in the bottom part of Fig.~\ref{fig_SADH}. Besides, a sigmoid layer is performed on the final feature map to constrain the predicted NWDs within (0, 1). 
	
	\subsubsection{Soft-style Binary Cross Entropy}
	For the learning of NWD, we propose a novel loss named SBCE for the localization regression in object detection, instead of utilizing the widely used $L_1$ loss. The SBCE could be formulated as: 
	\begin{equation}
	\label{eq_lossnwd}
	\mathcal{L}_{nwd} = -\sum_{i}^{} {log(1-\mid y_i - p_i \mid)},
	\end{equation}
	where $p_i$ is the predicted NWD of the $i$-th proposal, and $y_i$ is the label, i.e., the NWD between the $i$-th proposal and its corresponding ground truth. Referring to \cite{wang2021normalized}, $y_i$ is defined as:
	\begin{equation}
	y_i = NWD(B_i, G_i) = exp(-\frac{{W_2}(B_i, G_i)}{C}),
	\end{equation}
	where $C$ is a constant which is empirically set as the average size of acne lesions, $B_i$ and $G_i$ are the $i$-th proposal bounding box and its corresponding ground truth, respectively, and ${W_2}(B_i, G_i)$ is the Wasserstein coupling distance between them and could be calculated by:
	\begin{equation}
	\resizebox{.89\linewidth}{!}{$
		\displaystyle
		{W_2}(B, G)={\left \|{\left[ {{cx}_b, {cy}_b,\frac{w_b}{2},\frac{h_b}{2}} \right]}^T-{\left[{{cx}_g,{cy}_g,\frac{w_g}{2},\frac{h_g}{2}}\right]}^T\right \|}_2,
		$}
	\end{equation}
	where the $({cx}_b, {cy}_b)$, ${w_b}$, and ${h_b}$ denote the center coordinates, width and height of the proposal bounding box $B$ respectively, and the same is true for the ground truth $G$. 
	
	Referring to Eq.~\ref{eq_lossnwd}, it's interesting to find that Binary Cross Entropy is a special case of SBCE when given a hard label $y_i \in \{0, 1\}$, while SBCE enables the NWD prediction to be trained as regression task with soft label. Moreover, the logarithm function has a steeper gradient than $L_1$ loss, leading to a faster and better convergence.
	
	\begin{table*}
		\tbl{Comparison results with different generic object detection methods on AcneSCU \label{table_acnescu}}{
			\begin{threeparttable}
				\begin{tabular}{@{}lrrrrrrrr@{}}
					\toprule
					Method  & AP & $AP_s$ & $AP_m$ & $AP_l$ & $AR$ & $AR_s$ & $AR_m$ & $AR_l$ \\
					
					\botrule
					FCOS~\cite{tian2019fcos} & 0.247 & 0.197 & 0.298 & 0.320 & 0.649 & 0.411 & 0.714 & 0.791 \\		
					FSAF~\cite{zhu2019feature} & 0.402 & 0.332 & 0.445 & 0.680 & 0.805 & 0.806 & 0.813 & 0.750\\
					AutoAssign~\cite{tian2019fcos} & 0.471 & 0.451 & 0.509 & 0.409 & 0.873 & 0.862 & 0.876 & 0.850\\
					\botrule
					Faster R-CNN~\cite{ren2015faster} & 0.464 & 0.390 & 0.496 & 0.434 & 0.750 & 0.691 & 0.725 & 0.747\\
					Cascade R-CNN~\cite{cai2018cascade} & 0.451 & 0.416 & 0.497 & 0.463 & 0.766 & 0.788 & 0.748 & 0.688\\
					PANet~\cite{liu2018path} & 0.490 & 0.495 & 0.517 & 0.482 & 0.756 & 0.743 & 0.719 & 0.792\\
					Cascade Mask R-CNN~\cite{Cai_2019} & 0.455 & 0.480 & 0.477 & 0.437 & 0.708 & 0.745 & 0.661 & 0.696\\
					Mask Scoring R-CNN~\cite{huang2019mask} & 0.464 & 0.467 & 0.521 & 0.429 & 0.755 & 0.738 & 0.749 & 0.734\\
					HTC~\cite{chen2019hybrid} & 0.455 & 0.452 & 0.503 & 0.466 & 0.823 & 0.811 & 0.847 & 0.813\\
					GHM~\cite{li2019gradient} & 0.482 & 0.458 & 0.520 & 0.411 & 0.774 & 0.770 & 0.758 & 0.756\\
					Libra R-CNN~\cite{pang2021towards} & 0.477 & 0.475 & 0.523 & 0.446 & 0.784 & 0.768 & 0.759 & 0.786\\
					OHEM~\cite{shrivastava2016training} & 0.495 & 0.509 & 0.526 & 0.461 & 0.736 & 0.732 & 0.713 & 0.706\\
					PISA~\cite{cao2019prime} & 0.470 & 0.466 & 0.493 & 0.422 & 0.750 & 0.740 & 0.718 & 0.753\\
					\botrule
					Mask R-CNN~\cite{he2017mask} w/o masked crop & 0.457 & 0.354 & 0.496 & 0.425 & 0.729 & 0.625 & 0.709 & 0.727\\
					Mask R-CNN~\cite{he2017mask} & 0.481 & 0.488 & 0.515 & 0.469 & 0.750 & 0.742 & 0.714 & 0.759\\
					Mask R-CNN + SADH & 0.497 & 0.464 & 0.519 & 0.502 & 0.775 & 0.758 & 0.742 & 0.805\\
					Mask R-CNN + SADH + NWD & \textbf{0.507} & 0.494 & 0.555 & 0.472 & 0.775 & 0.761 & 0.754 & 0.786\\
					\botrule
				\end{tabular}
				\begin{tablenotes}
					\footnotesize
					\scriptsize{
						\item Please note all the methods are evaluated with masked crop unless specially marked with "w/o masked crop".
					}
				\end{tablenotes}
		\end{threeparttable}}
	\end{table*}
	
	\subsection{Training and Inferring}\label{subsec_traininfer}
	\subsubsection{Training} 
	In the training process, the overall loss of the proposed head structure is modified by adding the loss for NWD prediction, which could be formulated as:
	\begin{equation}
	\mathcal{L}_{rpn} = \mathcal{L}_{cls} + \mathcal{L}_{loc} + \lambda_{nwd}\times{\mathcal{L}_{nwd}},
	\end{equation}
	where $\mathcal{L}_{cls}$ and $\mathcal{L}_{loc}$ are the loss for classification and localization task in original RPN, respectively, and $\lambda_{nwd}$ is the hyper-parameter to adjust the weight of $\mathcal{L}_{nwd}$ and set as 1 by default. Same as the training of $\mathcal{L}_{loc}$, only the positive samples participate in training $\mathcal{L}_{nwd}$. 
	
	\subsubsection{Inferring}
	In the inferring process, the predicted NWD score is utilized to rectify the original classification score in RPN. To constrain the final classification confidence score $s_{final}$ within (0, 1), we formulated $s_{final}$ as:
	\begin{equation}
	s_{final} = \sqrt{{s^{(2-\omega_{nwd})}_{cls}} \times {p^{\omega_{nwd}}}},
	\end{equation}
	where $s_{cls}$ and $p$ denote the original classification score in RPN and the predicted NWD score, respectively, and $\omega_{nwd}$ is the hyper-parameter to adjust the weight of NWD score and set as 1 by default. 
	
	\section{Experiments}
	\subsection{Datasets.}
	We evaluate the proposed method on both the AcneSCU and ACNE04 datasets. For the AcneSCU dataset, we randomly split it into two parts, about $90\%$(248 images) are held out as the train set while the others are used as the test set. As some persons may exist in more than one image, all the persons in the test set are constrained not in the train set to avoid overfitting. For the ACNE04 dataset, it contains 1457 images with 18983 bounding boxes annotations of only one lesion category. Following \cite{wu2019joint}, the whole dataset is randomly split into $80\%$ train set and $20\%$ test set, consisting of 1165 and 292 images, respectively. 
	
	\subsection{Implementation}
	Our framework is built based on the Mask R-CNN implemented by MMDetection~\cite{mmdetection} with one NVIDIA GeForce 3090. The reasons for choosing Mask R-CNN as the baseline are: 1) FPN~\cite{lin2017feature} is utilized to enhance the detection of various-sized objects; 2) RPN~\cite{ren2015faster} is utilized for generating candidate regions efficiently, enabling the designing of more flexible and precise classifiers~\cite{hosang2015makes}; 3) it could take advantage of instance annotations to improve the detection performance~\cite{he2017mask}; 4) it's robust and widely utilized for comparison in generic object detection.
	
	We initialize the Mask R-CNN with parameters for COCO dataset in~\cite{he2017mask} and use a ResNet50 pretrained on ImageNet~\cite{deng2009imagenet} as the backbone. Deformable convolutional networks(DCN)~\cite{dai2017deformable} is added to enhance the transformation modeling capability. To output feature maps with suitable scales for acne lesions, only P2, P3, P4, and P5 feature maps of FPN~\cite{lin2017feature} are preserved. 2000 proposals per feature map from FPN are performed with an NMS with IoU threshold of 0.7, and 1000 proposals per sub-image are outputted at last to R-CNN. The anchors and proposals are assigned as positive samples if their IoUs with the matched ground truth are greater than 0.5 when training both RPN and RCNN. In the inferring process, an NMS with IoU threshold of 0.5 is utilized for all the detection bounding boxes, and only 200 per sub-image at most are outputted as detection results at last. In this paper, we find the overall performance works well with the hyper-parameter $\lambda_{nwd}$ and $\omega_{nwd}$ range within (0.8, 1.2), and they are both set as 1 by default. The network is trained by an SGD optimizer with 15 epochs, where the learning rate, momentum, and weight decay are 0.002, 0.9, and 0.0001, respectively. There are two ways to merge detection results of sub-images: 1) follow ~\cite{van2018you} to perform an NMS on the detection results of sub-images; 2) directly input the whole image into the model and set a larger number of proposals in RPN and RCNN. As the post-processing techniques are not the focus of this paper, without loss of generality, we only investigate the detection performance on sub-images.
	
	\subsection{Metrics}\label{subsec_metric}
	To efficiently evaluate the proposed methods, following \cite{min2021acnet}, we evaluate Average Precision(AP) and AR at IoU=0.5 with 100 detection boxes at most. Please note that IoU=0.5 is already a harsh condition in consideration of the arbitrary boundary and pixel-level human label error of small acne lesions; see Fig.~\ref{fig_dataset} for example. In addition, metrics with regard to different object sizes are also reported. Specifically, $AP$, $AP_s$, $AP_m$, and $AP_l$ denote the AP of all, small, medium, and large objects, respectively, and the same is true for AR. Please note AP is much more important than AR in object detection, because it evaluates the average precision under different recalls. However, AR is easy to improve by increasing the detection boxes. All this doesn’t conflict with our efforts to improve the quality and AR of the RPN proposals.
	
	\begin{tablehere}
		\tbl{Comparison results with previous state-of-the-art methods on ACNE04 \label{table_acne04}}{
			\begin{tabular}{@{}lr@{}}
				\toprule
				Method  & AP \\
				\botrule
				Faster R-CNN~\cite{ren2015faster} & 0.103  \\
				R-FCN~\cite{dai2016r} & 0.140  \\
				Rashataprucksa et al.~\cite{rashataprucksa2020acne} & 0.147  \\
				ACNet~\cite{min2021acnet} & 0.205  \\
				\botrule
				ours & \textbf{0.22} \\
				\botrule
		\end{tabular}}
	\end{tablehere}

\begin{table*}
\tbl{Results of our method with different plain detection frameworks on AcneSCU \label{table_generation}}{
	\begin{threeparttable}
		\begin{tabular}{@{}lrrrrrrrr@{}}
			\toprule
			Method  & AP & $AP_s$ & $AP_m$ & $AP_l$ & $AR$ & $AR_s$ & $AR_m$ & $AR_l$ \\
			\botrule
			HTC~\cite{chen2019hybrid} & 0.455 & 0.452 & 0.503 & 0.466 & 0.823 & 0.811 & 0.847 & 0.813\\
			HTC + SADH + NWD & \textbf{0.462} & 0.449 & 0.510 & 0.411 & 0.819 & 0.808 & 0.838 & 0.688\\
			\botrule
			Faster R-CNN~\cite{ren2015faster} & 0.464 & 0.390 & 0.496 & 0.434 & 0.750 & 0.691 & 0.725 & 0.747\\
			Faster R-CNN + SADH + NWD & \textbf{0.480} & 0.474 & 0.539 & 0.447 & 0.760 & 0.748 & 0.746 & 0.766\\
			\botrule
			Mask R-CNN~\cite{he2017mask} & 0.481 & 0.488 & 0.515 & 0.469 & 0.750 & 0.742 & 0.714 & 0.759\\
			Mask R-CNN + SADH + NWD & \textbf{0.507} & 0.494 & 0.555 & 0.472 & 0.775 & 0.761 & 0.754 & 0.786\\
			\botrule
		\end{tabular}
\end{threeparttable}}
\end{table*}

	\subsection{Results}
	To demonstrate the efficiency of the proposed method, state-of-the-art methods of both one-stage detection methods, e.g. FCOS~\cite{tian2019fcos}, FSAF~\cite{zhu2019feature}, and AutoAssign~\cite{tian2019fcos}, and the two-stage detection, e.g. Faster R-CNN~\cite{ren2015faster}, Cascade R-CNN~\cite{cai2018cascade}, PANet~\cite{liu2018path}, Cascade Mask R-CNN~\cite{Cai_2019}, Mask Scoring R-CNN~\cite{huang2019mask}, HTC~\cite{chen2019hybrid}, GHM~\cite{li2019gradient}, Libra R-CNN~\cite{pang2021towards}, OHEM~\cite{shrivastava2016training}, and PISA~\cite{cao2019prime}, are chosen as peer competitors on AcneSCU dataset. To make a fair comparison, we try our best to implement the compared methods with the same settings and hyper-parameters. For ACNE04, we compare the methods which have published results on it.
	
	Table~\ref{table_acnescu} shows the comparative results on AcneSCU. On the overall metric AP, our method outperforms all the compared state-of-the-art one-stage and two-stage generic object detection methods. Specifically, Mask R-CNN with masked crop outperforms the baseline with $2.4\%$ AP, which demonstrates the efficiency of the proposed masked crop. We argue that this is because masked crop could alleviate the influence caused by unavoidable partial lesions. Mask R-CNN with SADH achieves $1.6\%$ higher AP than that without SADH. This may benefit from a steeper classification confidence gradient, which could suppress the proposals of low IoUs. Compared to "Mask R-CNN + SADH", the NWD prediction branch could further improve the AP with $1.0\%$, which demonstrates the efficiency of NWD rectification, especially on medium and small lesions. In addition, we find AutoAssign achieves the greatest AR. That is because AutoAssign's detection number is much greater than that of our method, about 5:1. 
	
	Table~\ref{table_acne04} shows the comparative results on ACNE04. For a fair comparison,  we adopt the same data augmentation with previous method~\cite{min2021acnet}, which means our result doesn't benefit from the proposed masked crop data preprocessing method. In addition, we calculate the overall performance on the whole images to ensure comparison under the same evaluation condition. Despite without masked crop, the proposed outperforms all the previous methods on the public ACNE04 dataset.  
	
	Moreover, we also evaluate the generalization performance of our method with different detection frameworks on AcneSCU. As shown in Table~\ref{table_generation}, it could be found our method achieves consistent improvement with different detection frameworks, demonstrating the generalization performance of our method. Specifically, our method could improve HTC, Faster R-CNN, and Mask R-CNN with $0.7\%$, $1.6\%$, and $2.6\%$, respectively. The best improvement of Mask R-CNN may be due to its best detection performance of the baseline.
	
	\begin{tablehere}
		\tbl{Ablation study on ACNESCU \label{table_ablation}}{
			\begin{tabular}{ccccc}
				\toprule
				Group  & Convolution & Loss & Metric & AP\\
				\botrule
				\multirow{2}{*}{1} & $3\times3$ & - & - & 0.488 \\
				& $1\times1$ & - & - & \textbf{0.493} \\
				\botrule
				\multirow{3}{*}{2} & - & $L_1$ loss & - & 0.487 \\
				& - & SBCE & - & \textbf{0.498} \\
				\botrule
				\multirow{3}{*}{3} & - & - & IoU & 0.490 \\
				& - & - & GIoU & 0.480 \\
				& - & - & DIoU & 0.483 \\
				& - & - & NWD & \textbf{0.498} \\
				\botrule
		\end{tabular}}
	\end{tablehere}
	
	\begin{figure*}[!t]
		\centering{\includegraphics[width=0.90\textwidth]{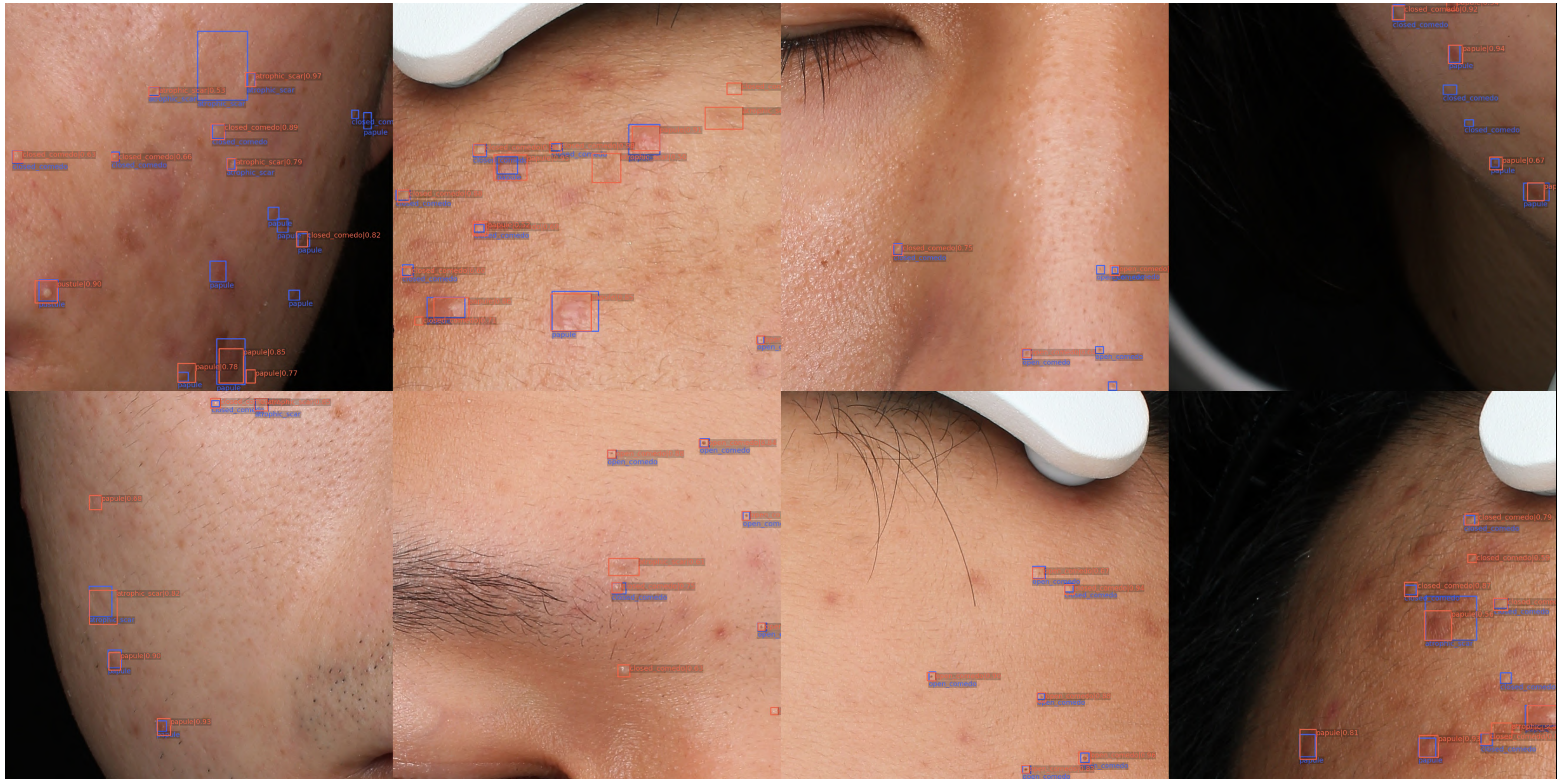}}
		\caption{The visual view of the detection results. Only the detected bounding boxes with confidence score greater than 0.5 are shown. The orange boxes indicate the detection results while the blue boxes indicate the ground truths.}
		\label{fig_visualview}
	\end{figure*}

	\subsection{Ablation Study}
	To validate different components of the proposed method, three groups of ablation experiments are performed, and the results are shown in Table~\ref{table_ablation}. Within each group, all the settings are same, including but not limited to DCNs in the backbone, GN in RPN head, and IoU threshold of NMS, except for the investigated component. Therefore, the influence of GN or DCNs on overall comparison is excluded. In addition, we try to make the smallest modification on the baseline. For example, we use the origin RPN for group 2 and 3. 
	
	\begin{figurehere}
		\centering{\includegraphics[width=0.41\textwidth]{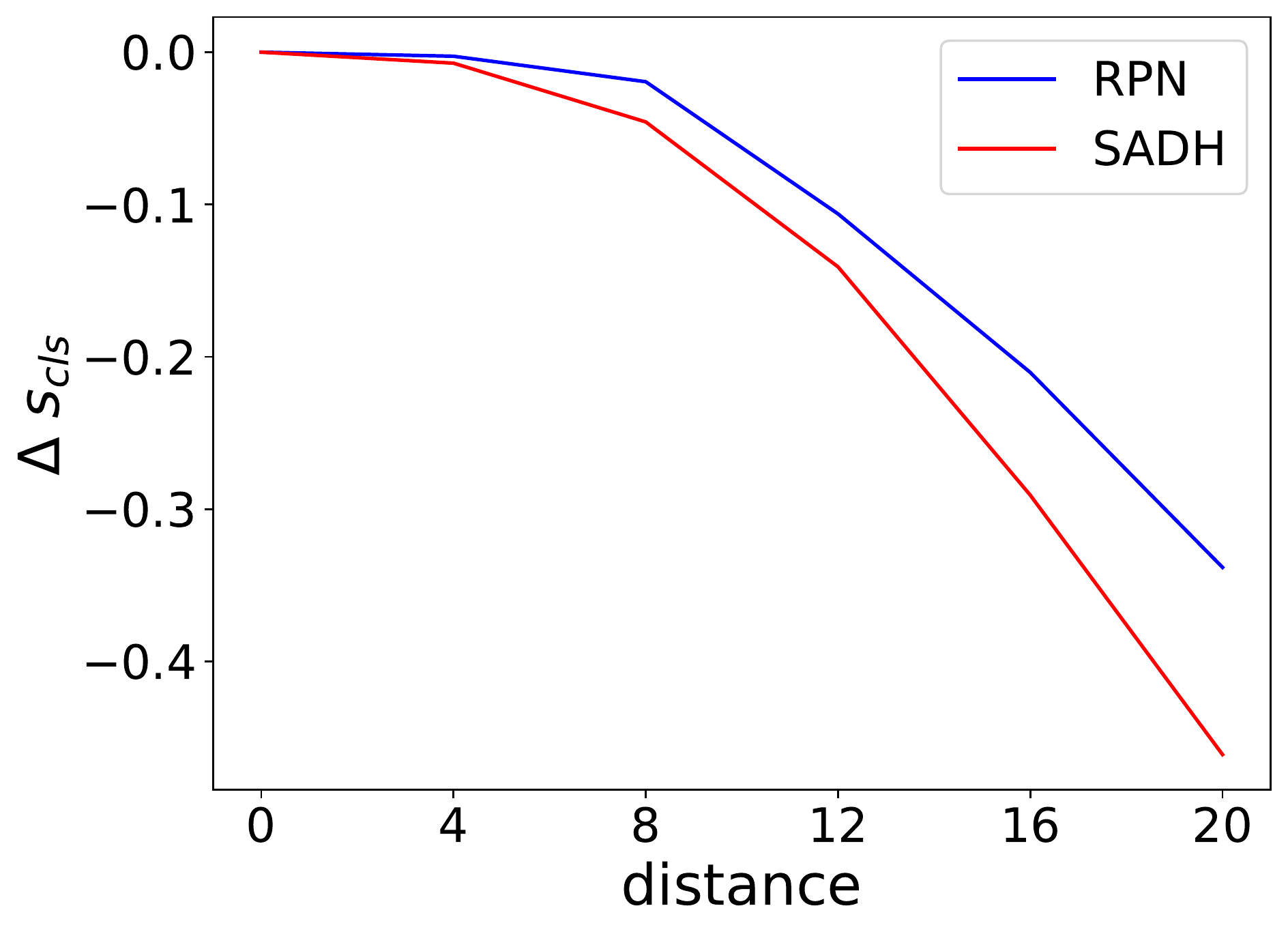}}
		\caption{The curve of the classification confidence gradient. The x-axis denotes the Manhattan distance between the feature map positions of the proposal and ground truth, while the y-axis denotes the classification confidence gradient, defined by $\Delta{s_{cls}} = s_{cls}(x)-s_{cls}(0)$, where $s_{cls}(x)$ denotes the classification score at the distance of x.}
		\label{fig_confidence_gradient}
	\end{figurehere}
	
	Specifically, the first group is conducted to analyze the kernel size of the convolution layer performed on the FPN feature map, emphasized with a bold font in Fig.~\ref{fig_SADH}, in the classification head of SADH. We find that \emph{$1\times1 convolution$} plays an important role in the success of SADH, which demonstrates that the kernel size indeed influences the independence of classification scores at adjacent positions on the final feature map. To valid the efficiency of SBCE loss, the second group compares the $L_1$ loss~\cite{girshick2015fast} and SBCE loss used in the NWD prediction branch. The results show that the proposed SBCE achieves $1.1\%$ higher AP than $L_1$ loss. We argue SBCE loss had a steeper curve than $L_1$ loss, leading to a faster and better convergence. To analysis the effectiveness of the localization confidence metric, the third group is performed by learning 4 different metrics, i.e., IoU, GIoU~\cite{rezatofighi2019generalized}, DIoU~\cite{zheng2020distance}, and NWD. It could be found that classification rectification with NWD is better than that with IoU, GIoU, and DIoU. This may be interpreted that NWD is a more suitable localization confidence metric for small acne lesions.

	\begin{figure*}[!t]
		\centering{\includegraphics[width=0.95\textwidth]{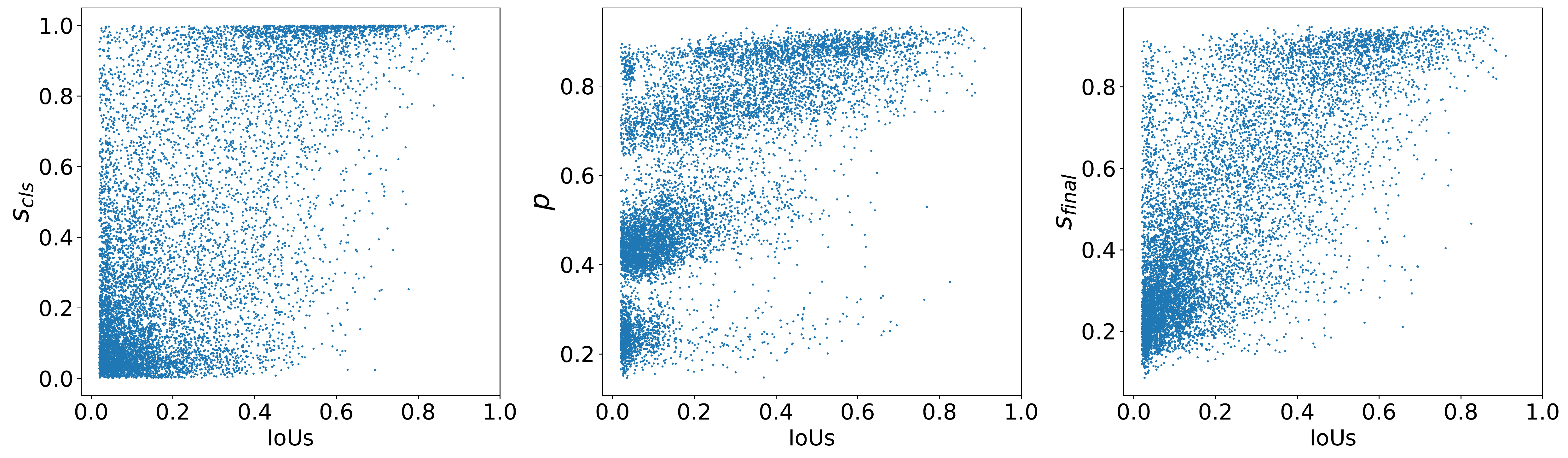}}
		\caption{Left: the correlation between the proposals' IoUs with the matched ground truths and the classification scores; Middle: the correlation between the IoUs and predicted NWDs; Right: the correlation between the IoUs and the rectified final scores. In summary, the Pearson correlation coefficients of the three groups are 0.63, 0.68, and 0.73, respectively.}
		\label{fig_nwd_iou_relation}
	\end{figure*}
	
	\section{Discussion}
	In this section, we firstly give some visual views to show the detection results of the proposed method in Subsection~\ref{subsec_visual_view}. Then, we intuitively to show the efficiency of the proposed method through in-depth analysis in Subsection~\ref{subsec_indepth}.

	\subsection{Visual View and Analysis of the Detection Results}\label{subsec_visual_view}
	The visual view of some detection results is shown in Fig.~\ref{fig_visualview}. It could be found most skin lesions, especially those which have obvious visual features, could be well detected. The reasons for false detection results could be concluded as: 1) some lesions located on cheeks do not have apparent texture features on the front-view facial images; 2) several close tiny are detected as one lesion; 3) some tiny lesions are too tiny(with sizes less than 20) to be detected for the harsh IoU threshold; 4) some acne lesions with arbitrary boundaries are hard to detect; 5) noisy labels. These problems would be the subjects of our future works.
	
	\subsection{In-depth Analysis}\label{subsec_indepth}
	To intuitively demonstrate the efficiency of the proposed SADH, we plot the steepness of the proposals' classification scores, which is referred to as \textbf{classification confidence gradient} in this paper, moving from the center of the ground truth to its boundary. As shown in Fig.~\ref{fig_confidence_gradient}, it could be found that SADH had a steeper classification confidence gradient than the vanilla RPN. We argue a steeper classification confidence gradient could generate more independent classification scores at adjacent positions on the final feature map, suppress the proposals with low IoUs but high classification scores, and then improve the detection of the hard samples.
	
	To demonstrate the effectiveness of the NWD prediction branch, we visualize the relations between the proposals' IoUs and classification scores $s_{cls}$, the predicted NWDs $p$, and the rectified final scores $s_{final}$, respectively. As shown in Fig.~\ref{fig_nwd_iou_relation}, there exists a large number of proposals with IoUs ranging from 0 to 0.4 and classification scores below 0.1. We argue that this is caused by the label assignment in RPN. The anchors whose IoUs below the label threshold, always set as 0.5, are regarded as negative samples. After the rectification of NWDs, these samples could be assigned with more appropriate classification confidences, i.e., the final scores $S_{final}$. As a result, the detection performance benefits from the higher relation between the classification confidences and IoUs. 
	
	\section{Conclusion}
	In this paper, we propose a new RPN head structure to improve the proposals' quality in acne detection. At first, a Spatial Aware Double Head structure is proposed to learn a steeper classification confidence gradient and suppress the redundant inferior proposals of easy samples. Then, an NWD prediction branch along with a Soft-style Binary Cross Entropy loss is proposed to rectify the proposals' classification scores. In addition, we construct a new dataset named AcneSCU, with high-resolution imageries, more precise annotations, and fine-grained lesion categories, to facilitate further research on acne detection. Extensive experiments on both AcneSCU and the public dataset ACNE04 demonstrate the efficiency of the proposed method.

	\nonumsection{Acknowledgments} \noindent This work was supported in part by Natural Science Foundation of China under Grant No.62025601.
	
	\renewcommand{\bibname}{References}
	\bibliographystyle{ws-ijns}
	\bibliography{reference}

\begin{thebibliography}{10}

\bibitem{bernardis2020development}
E.~Bernardis, H.~Shou, J.~S. Barbieri, P.~J. McMahon, M.~J. Perman, L.~A. Rola,
  J.~L. Streicher, J.~R. Treat, L.~Castelo-Soccio and A.~C. Yan, Development
  and initial validation of a multidimensional acne global grading system
  integrating primary lesions and secondary changes, {\em JAMA dermatology}
  {\bf 156}(3)  (2020)  296--302.

\bibitem{wu2019joint}
X.~Wu, N.~Wen, J.~Liang, Y.-K. Lai, D.~She, M.-M. Cheng and J.~Yang, Joint acne
  image grading and counting via label distribution learning, {\em Proceedings
  of the IEEE/CVF International Conference on Computer Vision\/}, 2019, pp.
  10642--10651.

\bibitem{barbieri2021patient}
J.~S. Barbieri, R.~Fulton, R.~Neergaard, M.~N. Nelson, F.~K. Barg and D.~J.
  Margolis, Patient perspectives on the lived experience of acne and its
  treatment among adult women with acne: A qualitative study, {\em JAMA
  dermatology} {\bf 157}(9)  (2021)  1040--1046.

\bibitem{kittigul2016automatic}
N.~Kittigul and B.~Uyyanonvara, Automatic acne detection system for medical
  treatment progress report, {\em 2016 7th International Conference of
  Information and Communication Technology for Embedded Systems (IC-ICTES)\/},
  IEEE2016, pp. 41--44.

\bibitem{maroni2017automated}
G.~Maroni, M.~Ermidoro, F.~Previdi and G.~Bigini, Automated detection,
  extraction and counting of acne lesions for automatic evaluation and tracking
  of acne severity, {\em 2017 IEEE symposium series on computational
  intelligence (SSCI)\/},  IEEE2017, pp. 1--6.

\bibitem{rashataprucksa2020acne}
K.~Rashataprucksa, C.~Chuangchaichatchavarn, S.~Triukose, S.~Nitinawarat,
  M.~Pongprutthipan and K.~Piromsopa, Acne detection with deep neural networks,
  {\em 2020 2nd International Conference on Image Processing and Machine
  Vision\/}, 2020, pp. 53--56.

\bibitem{min2021acnet}
K.~Min, G.-H. Lee and S.-W. Lee, Acnet: Mask-aware attention with dynamic
  context enhancement for robust acne detection, {\em arXiv preprint
  arXiv:2105.14891}   (2021).

\bibitem{ren2015faster}
S.~Ren, K.~He, R.~Girshick and J.~Sun, Faster r-cnn: Towards real-time object
  detection with region proposal networks, {\em Advances in neural information
  processing systems} {\bf 28}  (2015)  91--99.

\bibitem{girshick2014rich}
R.~Girshick, J.~Donahue, T.~Darrell and J.~Malik, Rich feature hierarchies for
  accurate object detection and semantic segmentation, {\em Proceedings of the
  IEEE conference on computer vision and pattern recognition\/}, 2014, pp.
  580--587.

\bibitem{hosang2015makes}
J.~Hosang, R.~Benenson, P.~Doll{\'a}r and B.~Schiele, What makes for effective
  detection proposals?, {\em IEEE transactions on pattern analysis and machine
  intelligence} {\bf 38}(4)  (2015)  814--830.

\bibitem{zou2019object}
Z.~Zou, Z.~Shi, Y.~Guo and J.~Ye, Object detection in 20 years: A survey, {\em
  arXiv preprint arXiv:1905.05055}   (2019).

\bibitem{wang2021normalized}
J.~Wang, C.~Xu, W.~Yang and L.~Yu, A normalized gaussian wasserstein distance
  for tiny object detection, {\em arXiv preprint arXiv:2110.13389}   (2021).

\bibitem{long2015fully}
J.~Long, E.~Shelhamer and T.~Darrell, Fully convolutional networks for semantic
  segmentation, {\em Proceedings of the IEEE conference on computer vision and
  pattern recognition\/}, 2015, pp. 3431--3440.

\bibitem{law2018cornernet}
H.~Law and J.~Deng, Cornernet: Detecting objects as paired keypoints, {\em
  Proceedings of the European conference on computer vision (ECCV)\/}, 2018,
  pp. 734--750.

\bibitem{duan2019centernet}
K.~Duan, S.~Bai, L.~Xie, H.~Qi, Q.~Huang and Q.~Tian, Centernet: Keypoint
  triplets for object detection, {\em Proceedings of the IEEE/CVF international
  conference on computer vision\/}, 2019, pp. 6569--6578.

\bibitem{ma2021anchor}
T.~Ma, W.~Tian, P.~Kuang and Y.~Xie, An anchor-free object detector with novel
  corner matching method, {\em Knowledge-Based Systems} {\bf 224}  (2021) p.
  107083.

\bibitem{sun2020exploring}
X.~Sun, C.~Chen, J.~Dong, D.~Liu and G.~Hu, Exploring ubiquitous relations for
  boosting classification and localization, {\em Knowledge-Based Systems} {\bf
  196}  (2020) p. 105824.

\bibitem{he2017mask}
K.~He, G.~Gkioxari, P.~Doll{\'a}r and R.~Girshick, Mask r-cnn, {\em Proceedings
  of the IEEE international conference on computer vision\/}, 2017, pp.
  2961--2969.

\bibitem{wu2020rethinking}
Y.~Wu, Y.~Chen, L.~Yuan, Z.~Liu, L.~Wang, H.~Li and Y.~Fu, Rethinking
  classification and localization for object detection, {\em Proceedings of the
  IEEE/CVF conference on computer vision and pattern recognition\/}, 2020, pp.
  10186--10195.

\bibitem{tian2019fcos}
Z.~Tian, C.~Shen, H.~Chen and T.~He, Fcos: Fully convolutional one-stage object
  detection, {\em Proceedings of the IEEE/CVF international conference on
  computer vision\/}, 2019, pp. 9627--9636.

\bibitem{jiang2018acquisition}
B.~Jiang, R.~Luo, J.~Mao, T.~Xiao and Y.~Jiang, Acquisition of localization
  confidence for accurate object detection, {\em Proceedings of the European
  conference on computer vision (ECCV)\/}, 2018, pp. 784--799.

\bibitem{huang2019mask}
Z.~Huang, L.~Huang, Y.~Gong, C.~Huang and X.~Wang, Mask scoring r-cnn, {\em
  Proceedings of the IEEE/CVF Conference on Computer Vision and Pattern
  Recognition\/}, 2019, pp. 6409--6418.

\bibitem{zhu2020autoassign}
B.~Zhu, J.~Wang, Z.~Jiang, F.~Zong, S.~Liu, Z.~Li and J.~Sun, Autoassign:
  Differentiable label assignment for dense object detection, {\em arXiv
  preprint arXiv:2007.03496}   (2020).

\bibitem{zhu2019feature}
C.~Zhu, Y.~He and M.~Savvides, Feature selective anchor-free module for
  single-shot object detection, {\em Proceedings of the IEEE Conference on
  Computer Vision and Pattern Recognition\/}, 2019, pp. 840--849.

\bibitem{cai2018cascade}
Z.~Cai and N.~Vasconcelos, Cascade r-cnn: Delving into high quality object
  detection, {\em Proceedings of the IEEE conference on computer vision and
  pattern recognition\/}, 2018, pp. 6154--6162.

\bibitem{liu2018path}
S.~Liu, L.~Qi, H.~Qin, J.~Shi and J.~Jia, Path aggregation network for instance
  segmentation, {\em Proceedings of the IEEE conference on computer vision and
  pattern recognition\/}, 2018, pp. 8759--8768.

\bibitem{Cai_2019}
Z.~Cai and N.~Vasconcelos, Cascade r-cnn: High quality object detection and
  instance segmentation, {\em IEEE Transactions on Pattern Analysis and Machine
  Intelligence}   (2019) p. 1–1.

\bibitem{chen2019hybrid}
K.~Chen, J.~Pang, J.~Wang, Y.~Xiong, X.~Li, S.~Sun, W.~Feng, Z.~Liu, J.~Shi,
  W.~Ouyang, C.~C. Loy and D.~Lin, Hybrid task cascade for instance
  segmentation, {\em IEEE Conference on Computer Vision and Pattern
  Recognition\/}, 2019.

\bibitem{li2019gradient}
B.~Li, Y.~Liu and X.~Wang, Gradient harmonized single-stage detector, {\em AAAI
  Conference on Artificial Intelligence\/}, 2019.

\bibitem{pang2021towards}
J.~Pang, K.~Chen, Q.~Li, Z.~Xu, H.~Feng, J.~Shi, W.~Ouyang and D.~Lin, Towards
  balanced learning for instance recognition, {\em International Journal of
  Computer Vision} {\bf 129}(5)  (2021)  1376--1393.

\bibitem{shrivastava2016training}
A.~Shrivastava, A.~Gupta and R.~Girshick, Training region-based object
  detectors with online hard example mining, {\em Proceedings of the IEEE
  conference on computer vision and pattern recognition\/}, 2016, pp. 761--769.

\bibitem{cao2019prime}
Y.~Cao, K.~Chen, C.~C. Loy and D.~Lin, Prime sample attention in object
  detection, {\em IEEE Conference on Computer Vision and Pattern
  Recognition\/}, 2020.

\bibitem{mmdetection}
K.~Chen, J.~Wang, J.~Pang, Y.~Cao, Y.~Xiong, X.~Li, S.~Sun, W.~Feng, Z.~Liu,
  J.~Xu, Z.~Zhang, D.~Cheng, C.~Zhu, T.~Cheng, Q.~Zhao, B.~Li, X.~Lu, R.~Zhu,
  Y.~Wu, J.~Dai, J.~Wang, J.~Shi, W.~Ouyang, C.~C. Loy and D.~Lin,
  {MMDetection}: Open mmlab detection toolbox and benchmark, {\em arXiv
  preprint arXiv:1906.07155}   (2019).

\bibitem{lin2017feature}
T.-Y. Lin, P.~Doll{\'a}r, R.~Girshick, K.~He, B.~Hariharan and S.~Belongie,
  Feature pyramid networks for object detection, {\em Proceedings of the IEEE
  conference on computer vision and pattern recognition\/}, 2017, pp.
  2117--2125.

\bibitem{deng2009imagenet}
J.~Deng, W.~Dong, R.~Socher, L.-J. Li, K.~Li and L.~Fei-Fei, Imagenet: A
  large-scale hierarchical image database, {\em 2009 IEEE conference on
  computer vision and pattern recognition\/},  Ieee2009, pp. 248--255.

\bibitem{dai2017deformable}
J.~Dai, H.~Qi, Y.~Xiong, Y.~Li, G.~Zhang, H.~Hu and Y.~Wei, Deformable
  convolutional networks, {\em Proceedings of the IEEE international conference
  on computer vision\/}, 2017, pp. 764--773.

\bibitem{van2018you}
A.~Van~Etten, You only look twice: Rapid multi-scale object detection in
  satellite imagery, {\em arXiv preprint arXiv:1805.09512}   (2018).

\bibitem{dai2016r}
J.~Dai, Y.~Li, K.~He and J.~Sun, R-fcn: Object detection via region-based fully
  convolutional networks, {\em Advances in neural information processing
  systems\/}, 2016, pp. 379--387.

\bibitem{girshick2015fast}
R.~Girshick, Fast r-cnn, {\em Proceedings of the IEEE international conference
  on computer vision\/}, 2015, pp. 1440--1448.

\bibitem{rezatofighi2019generalized}
H.~Rezatofighi, N.~Tsoi, J.~Gwak, A.~Sadeghian, I.~Reid and S.~Savarese,
  Generalized intersection over union: A metric and a loss for bounding box
  regression, {\em Proceedings of the IEEE/CVF Conference on Computer Vision
  and Pattern Recognition\/}, 2019, pp. 658--666.

\bibitem{zheng2020distance}
Z.~Zheng, P.~Wang, W.~Liu, J.~Li, R.~Ye and D.~Ren, Distance-iou loss: Faster
  and better learning for bounding box regression, {\em Proceedings of the AAAI
  Conference on Artificial Intelligence\/},   {\bf 34}(07)2020, pp.
  12993--13000.

\end{thebibliography}
\end{multicols}

\end{document}